\documentclass[a4paper, 10 pt, conference]{ieeeconf}  

\IEEEoverridecommandlockouts                              

\usepackage{graphicx}      
\usepackage{pgf}      
\usepackage{hyperref} 
\usepackage{subfig}
\usepackage{gensymb}

\usepackage{cite}
\usepackage{mathtools}
\usepackage{bm}
\usepackage{stfloats}
\usepackage{amsmath,amssymb,amsfonts}
\usepackage{algorithmic}
\usepackage{textcomp}

\usepackage{graphics}

\def\BibTeX{{\rm B\kern-.05em{\sc i\kern-.025em b}\kern-.08em
    T\kern-.1667em\lower.7ex\hbox{E}\kern-.125emX}}

\begin{document}

\title{Combined Aerial Cooperative Tethered Carrying and Path Planning for Quadrotors in Confined Environments}

\author{Marios-Nektarios Stamatopoulos$^{1}$, Panagiotis Koustoumpardis$^{2}$, Achilleas Seisa$^{1}$ and George Nikolakopoulos$^{1}$
\thanks{$^{1}$Robotics and Artificial Intelligence Group, Department of Computer, Electrical and Space Engineering, Lule\aa\,\,University of Technology, Lule\aa\,\,Sweden}%
\thanks{$^{2}$ Robotics Group, Department of Mechanical Engineering and
Aeronautics, University of Patras, Patras, Greece
}%
\thanks{Corresponding author's mail: \texttt{marsta@ltu.se} }
}
\maketitle

\begin{abstract}
In this article, a novel combined aerial cooperative tethered carrying and path planning framework is introduced with a special focus on applications in confined environments. The proposed work is aiming towards solving the path planning problem for the formation of two quadrotors, while having a rope hanging below them and passing through or around obstacles. A novel composition mechanism is proposed, which simplifies the degrees of freedom of the combined aerial system and expresses the corresponding states in a compact form. Given the state of the composition, a dynamic body is generated that encapsulates the quadrotors-rope system and makes the procedure of collision checking between the system and the environment more efficient. By utilizing the above two abstractions, an RRT path planning scheme is implemented and a collision-free path for the formation is generated. This path is decomposed back to the quadrotors’ desired positions that are fed to the Model Predictive Controller (MPC) for each one. The efficiency of the proposed framework is experimentally evaluated.
\end{abstract}

\section{INTRODUCTION}

In recent years, the utilization of quadrotors has become popular for various load-carrying applications \cite{loadTransportSurvey, CatenaryLoadPulling}, while one of the most common has been the last mile delivery \cite{lastMile}, where many investigations have been made for the payload transportation, mainly due to their benefit of being able to move and maneuver fast and precisely in confined spaces. However, their short battery life, in combination with the limited actuation power, makes them unable in some cases to carry and manipulate bigger and heavier payloads. Thus, the need of having two or more quadrotors collaborating and carrying the payload together, arises. A usual approach for solving this problem, is having the payload suspended from a deformable object (e.g. rope) connecting to the Unmanned Aerial Vehicles (UAVs) \cite{payloadCarryingExample, payloadCarryingExample2}. 

The problem of the UAVs' collaboration for sharing the payload has been introduced in many studies. Towards the transportation of a flexible hose, \cite{Kotaru2020} modeled it as a series of smaller discrete links and by deriving coordinate-free dynamics showed the differential-flatness of this under-actuated system. In \cite{pereira2017}, the quadrotors-load system is transformed into three decoupled sub-systems concerning the position of the load, the angle between the cables and the plane formed by the cables, with the two latter having double-integrator dynamics, while not taking into account the curve of each rope though. Then, controllers from the literature, similar to those of an under-actuated aerial vehicle, were used to control the sub-systems. Researchers have also introduced a hybrid modeling of the object based on catenaries and parabolas, depending on the distance between the UAVs and a follow-the-leader formation in, \cite{hybrid-rope-modeling}. 

In \cite{decentralized_transport_of_fabrics}, a fabric, approximated by cables with a uniformly distributed weight, was modeled based on catenary curves and led to a system of springs and dumpers creating forces on each UAV and a fully decentralized control was implemented. Having the limitation that the quadrotors maintain the same altitude at all the time. A different approach is followed in \cite{Catenary_robot}, where a centralized scheme is introduced, aiming towards the position, orientation, and span control of the rope that is also modeled as a catenary curve. The catenary is consisted of five states (3 for position, yaw, span) and all of them are controlled by changing the position and the orientation of the two quadrotors. It is also assumed that the robots move slowly enough to not cause any swing of the cable hanging below. A geometrical controller was utilized to command the UAVs’ positions and to track the desired catenary trajectory, while the limitation that the same altitude was still a demand. Aiming to continuously connect the UAVs with a power supply, \cite{obstacleAvoidance} calculated the shape of the connecting cable using catenary curves. Two static strategies (either horizontal or vertical) for avoiding collisions with objects of the environment were proposed by only changing the altitude of the lowest point of the cable in the limited area. However, in this case the combination of the above was not implemented and the obstacle avoidance, despite being fast and simple, was usable in the case of having one end of the cable hanging from a stationary prop. 

\begin{figure}[htbp]
    \begin{center}
        \includegraphics[width=0.74\columnwidth]{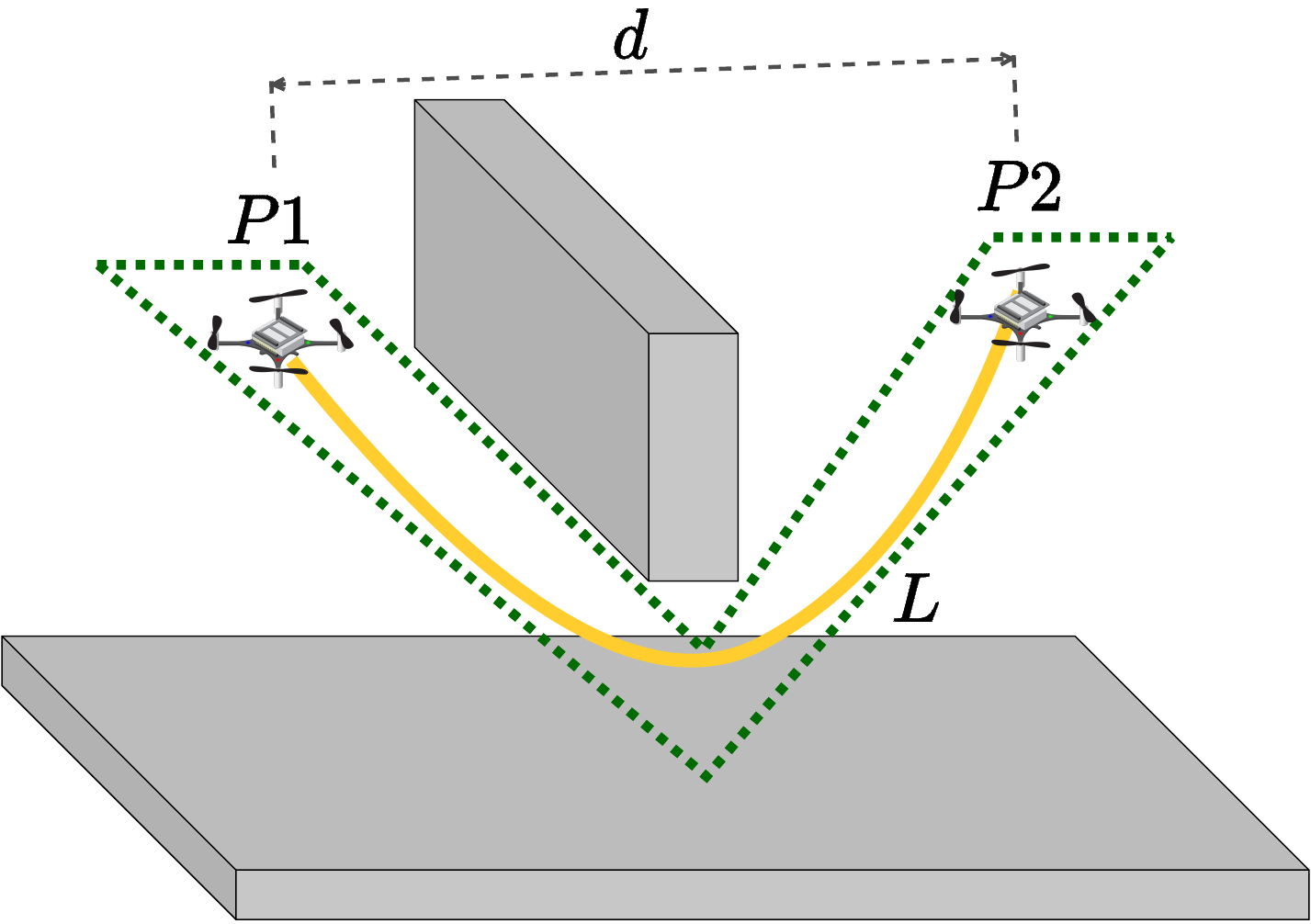}    
        \caption{Concept Representation. The formation is avoiding an obstacle while maintaining no-collision between the dynamic rigid body (green) and the obstacles.}
        \label{fig:concept}
    \end{center}
\end{figure}


Based on the current literature and to the best of the authors' knowledge, the majority of researchers are focusing on implementing control laws, waypoint, and trajectory tracking for the formation of the overall system. Alternatively to this, in this article, a path planning scheme for the UAVs and rope formation is introduced, aiming to generate and execute collision-free paths as depicted in Fig. \ref{fig:concept}.

\begin{figure*}[b]
	\centering
	\includegraphics[width=0.94\textwidth]{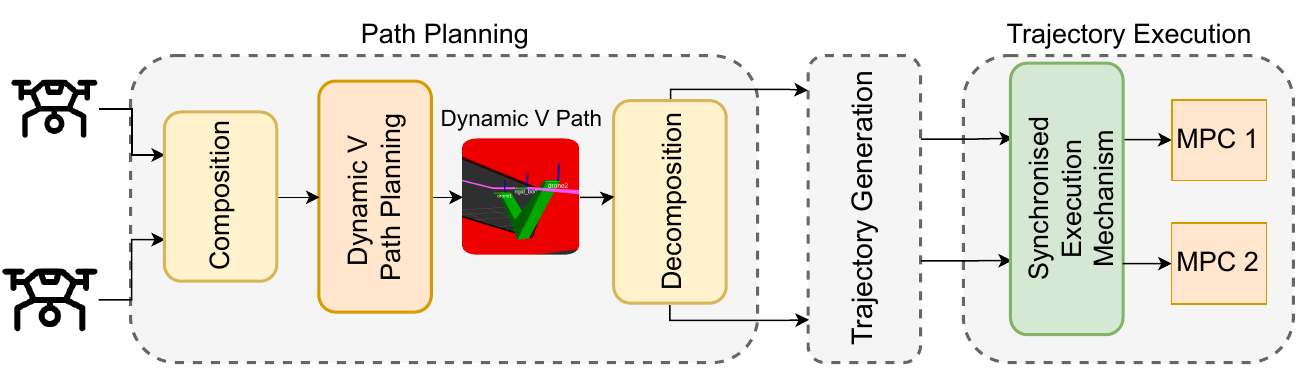}
  	\caption{System Block Diagram}
  	\label{fig:block_diagram}
\end{figure*}

This approach addresses the problem of autonomous transportation of hoses, fabrics and tethered systems, into indoor confined spaces with the presence of obstacles with small openings and holes like tunnels or damaged buildings. Thus, the contributions of the article can be stated as it follows. Initially, a novel composition mechanism is introduced that is able to narrow down the degrees of freedom to six from eight. In this way, all the possible states and combinations of the path planning scheme can be expressed in a compact and precise form. Moreover, a novel dynamic rigid body mechanism is introduced that encloses the formation of the UAVs for its different states. It simplifies and makes the collision-checking process with the surrounding environment faster. In addition, the dynamic shape, along with the formation transformation and its inverse transformation, leads to the reduction and abstraction of the model. Thus, making it possible to implement a path planning scheme from a starting to a goal formation state, while maintaining a successful lift. This is done by maneuvering around or inside obstacles, by changing the shape of the quadrotors-rope system. Finally, the path of the formation and its states are decomposed back to the two UAVs' positions, forming waypoints in space, where each UAV has to pass through. The block diagram of the proposed system can be seen in Fig. \ref{fig:block_diagram}, while the overall components will be analyzed.


The rest of the article is structured as it follows. In Section 2, the rope modelling is established, and in Section 3 the introduced novel composition and decomposition approach for way point generation is analyzed. In Section 4 the dynamic rigid body is formulated, and it is coupled with the corresponding path planner scheme, while in Section 5 the UAV control framework is described. Finally, in Section 6 experimental results to evaluate the efficiency of the framework are presented, followed by the conclusions in Section 8.

\section{Rope Modelling}
The UAVs are connected with each other via a rope that is attached to their bottom part, while the modeling of the rope is based on catenary curves. Prior to the calculation of the $3D$ curve of the catenary, the plane passing through the two UAVs needs to be extracted, and a $2D$ curve is calculated, as shown in Fig. \ref{fig:composition_modelling}.a. The 2D curve is expressed in Eq.~\ref{eq:cat_equation}.

\begin{equation}\label{eq:cat_equation}
f(x)=a \cdot cosh(\frac{x-b}{a}) + c
\end{equation}

$a\in \mathbb{R}$ is a scaling factor, $b \in \mathbb{R}$ is the center of the curve, and $c \in \mathbb{R}$ is the vertical offset from the origin. These three parameters are extracted by fulfilling the constraints of Eq.~\ref{eq:rope_model_pos_conditions}.

\begin{equation} \label{eq:rope_model_pos_conditions}
\begin{gathered}
f(x_1) = y1 \\
f(x2) = y2 \\
 \int_{x1}^{x2} \sqrt{1+ (f'(x))^2} \,dx =L
\end{gathered}
\end{equation}

$(x_1, y_1)$ and $(x_2, y_2)$ are the 2D positions of the UAVs and $L\in \mathbb{R}^+$ is the length of the rope. 
Assuming that $\overline{x}= \frac{x_1+x_2}{2}$,$A=\frac{dx}{2a},B=\frac{\overline{x}-b}{a}$, $r = \frac{ \sqrt{L^2 -dy^2} }{dx}$ the problem of calculating $a,b,c$ is simplified into solving Eq.~\ref{eq:r=sinh} for $A$.

\begin{equation}\label{eq:r=sinh}
r = \frac{sinh(A)}{A}
\end{equation}

Eq.~\ref{eq:r=sinh} is numerically solved by applying Newton's method, leading to the iterative Eq.~\ref{eq:Ai_iter}.

\begin{equation}\label{eq:Ai_iter}
A_{n+1} = A_n - \frac{sinh(A_ n) - rA_n}{cosh(A_n) - r}
\end{equation}

After solving for $A$, the initial parameters can be calculated in Eq.~\ref{eq:paramExtraction}.

\begin{equation} \label{eq:paramExtraction}
\begin{gathered}
a = \frac{dx}{2A} \\
b = {x} - a \cdot arctan(\frac{dy}{L}) \\ 
c = y_i - a\cdot cosh(\frac{x_i-b}{a})  (i= 1,2)
\end{gathered}
\end{equation}

Once the $2D$ curve is calculated, it is transformed back to the 3D space by utilizing the initial transformation of the plane.

\section{Composition - Decomposition} \label{sec:composition_decomposition}
\begin{figure}[ht]
\begin{center}
    \subfloat[Side View]{{\includegraphics[width=6.5cm]{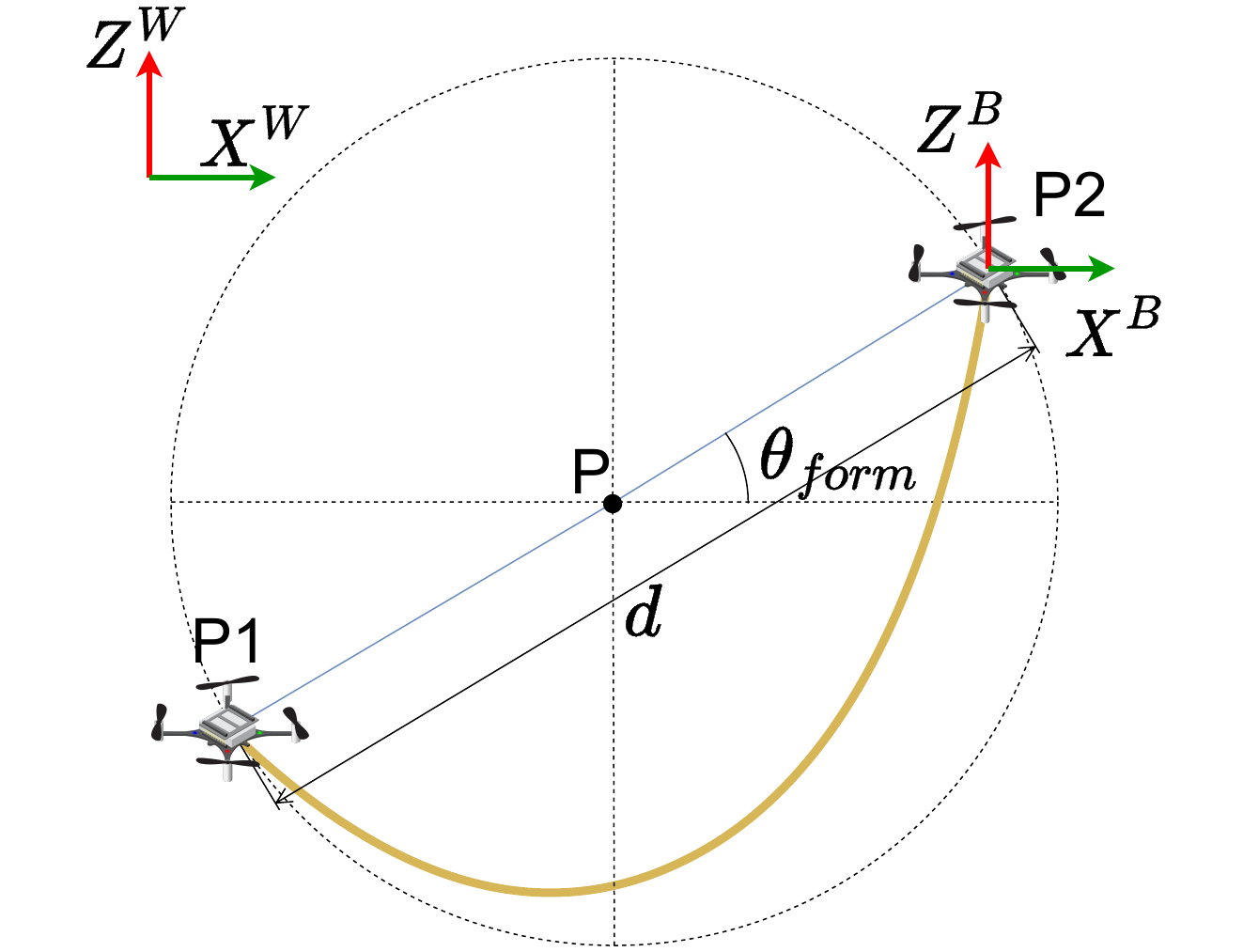} }}
    \qquad
    \subfloat[Top View]{{\includegraphics[width=6.5cm]{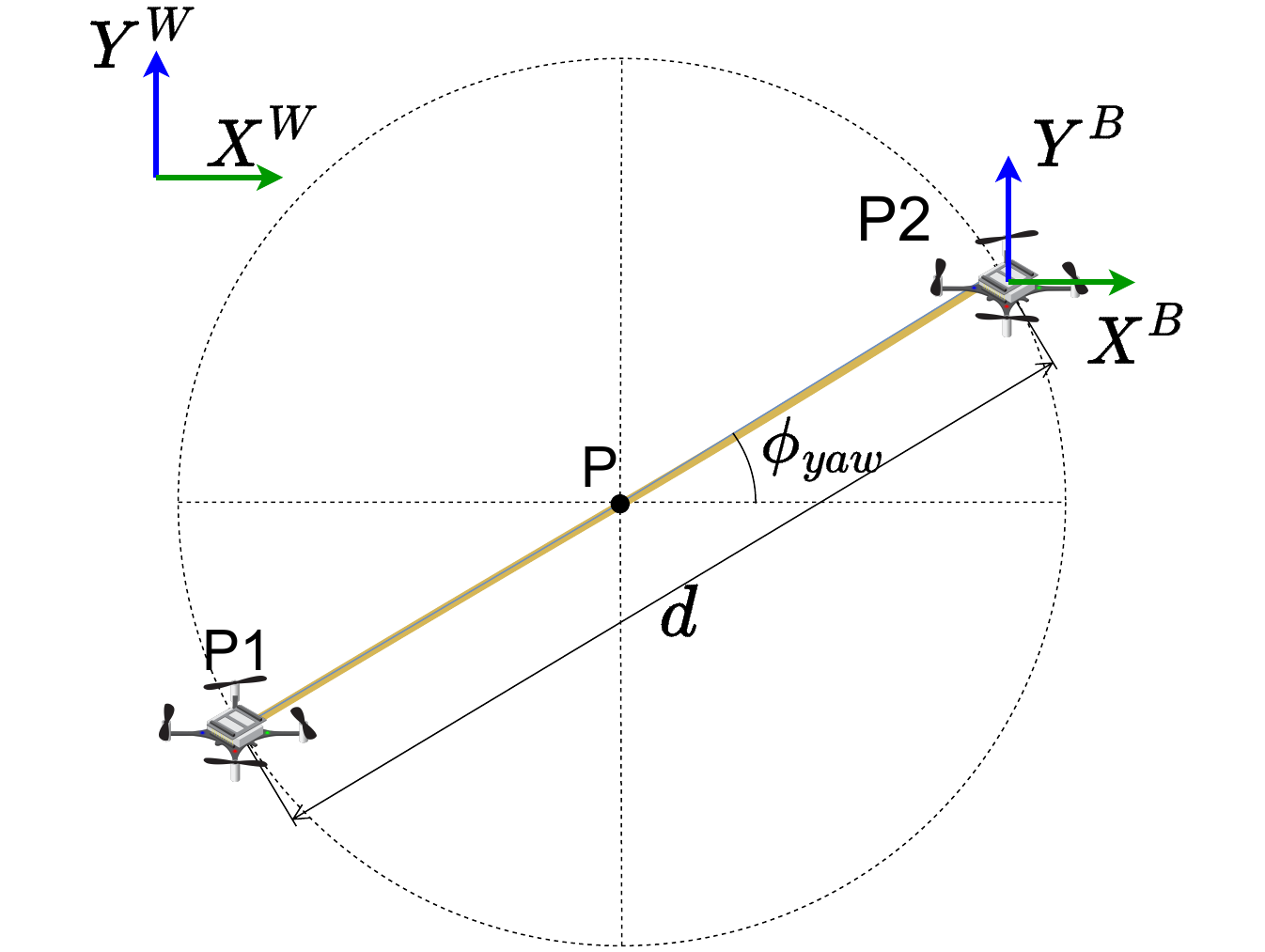} }}
    
    \caption{Composition Modelling. Side View (a) and Top View (b) of the plane shows how $\theta_{form}, \ d$ and $\phi_{yaw}$ are calculated.}
    \label{fig:composition_modelling}
\end{center}
    
\end{figure}

To reduce the dimensions of the search space, a transformation is implemented which transforms the two independent state vectors (one for each UAV) of the 4D space ($x,y,z$ and $yaw$) into a common one. The roll and pitch angles of the quadrotor are deemed constant since they do not provide any extra versatility to the system. The outcome of the transformation is a dynamically changing formation that can fully describe all the possible combinations of the two UAVs manipulating the rope attached to them, while maintaining no collision between the UAVs. The rope is kept under no tension, and untangled. Essentially, a $2D$ plane is calculated by the $3D$ positions of the UAVs and then, the distance between them and their relative angle are calculated. An illustration is shown in Fig. \ref{fig:composition_modelling}.

\subsection{Formation Composition}
The transformation is calculated, by knowing the position of the two UAVs ($\vec{P1} \in \mathbb{R}^3$ and $\vec{P2} \in \mathbb{R}^3$) in space. The new coordinates $\vec{P} \in \mathbb{R}^3$ of the composition are the middle of the position of the two UAVs. Then, the $2D$ plane perpendicular to the $X-Y$ plane is calculated, and the $\phi_{yaw} \in \mathbb{R}$ angle is formed by the plane and the $X$-axis. Furthermore, the $2D$ Euclidean distance between the UAVs on the plane is the distance $d \in \mathbb{R}^+$ and their relative elevation is expressed by the angle $\theta_{form} \in [-60\degree ,60\degree]$ formed by the line connecting the UAVs and the $X$-axis of the plane. The corresponding expressions are presented in Eq.~\ref{eq:test}.

\begin{equation} \label{eq:test}
\begin{gathered}
\vec{P} = \frac{ \vec{P1} +\vec{P2} }{2} \\
\phi_{yaw} = arctan(\frac{dy}{dx}) \\ 
d = \sqrt{dx^2+dy^2} \\ 
\theta_{form} = arctan(\frac{dz}{dx}) \\ 
\end{gathered}
\end{equation}

\subsection{Formation Decomposition}
Similarly, knowing the $\phi_{yaw}$ angle, the origin $\vec{P}$ of the plane and both UAVs' distance $d$ and angle $\theta_{form}$, the position of each UAV $\vec{P1},\vec{P2}$ can be extracted in Eq.~\ref{eq:test2}. Firstly, the projection of each UAV is calculated on the $2D$ plane.
\begin{equation} \label{eq:test2}
\begin{gathered}
R= \frac{d}{2}\\
\vec{P1}_{2D} =( +R\cos(\theta_{form}),+R\sin(\theta_{form}) ) \\
\vec{P2}_{2D} =( -R\cos(\theta_{form}),-R\sin(\theta_{form}) ) \\
\end{gathered}
\end{equation}

So knowing the position on the $2D$ plane and the transformation of the plane, the $3D$ position can be extracted easily by taking the inverse transformation $M$ of the plane.

\begin{equation}
M =
\begin{bmatrix}
cos(\phi_{yaw}) & -sin(\phi_{yaw}) & 0 & P_x \\
sin(\phi_{yaw}) & cos(\phi_{yaw}) & 0 & P_y \\
0 & 0 & 1 & P_z \\
0 & 0 & 0 & 1 \\
\end{bmatrix}
\end{equation}

\section{Path Planning}
A robust, precise and fast collision checking between the obstacles and quadrotors-rope system should be implemented. Hence, having as input one state vector of this transformation $[x,y,z, \phi_{yaw}, d, \theta_{form}]$, a rigid body is calculated that encloses both the UAVs and the whole rope, so the collision checking of the whole complex system is simplified into the simple collision check between this rigid body and the obstacle. This body is a set of 3D points that need to be calculated, and after applying a predefined triangulation method, they lead to a 3D mesh. The form of this mesh is shown in Eq. \ref{eq:VbodyDefinition}.

\begin{multline}\label{eq:VbodyDefinition}
\mathbf{V} = 
[\begin{matrix} P_{lower} & P_{innerR} & P_{right} & P_{intersect} & \end{matrix} \\
 \begin{matrix} P_{left} & P_{innerL} & P^{off}_{lower} & \dots & P^{off}_{innerL} \end{matrix}]
\end{multline}

\begin{figure}[htbp]
\begin{center}
    \subfloat[$2D$ Calculation]{{\includegraphics[width=0.86\columnwidth]{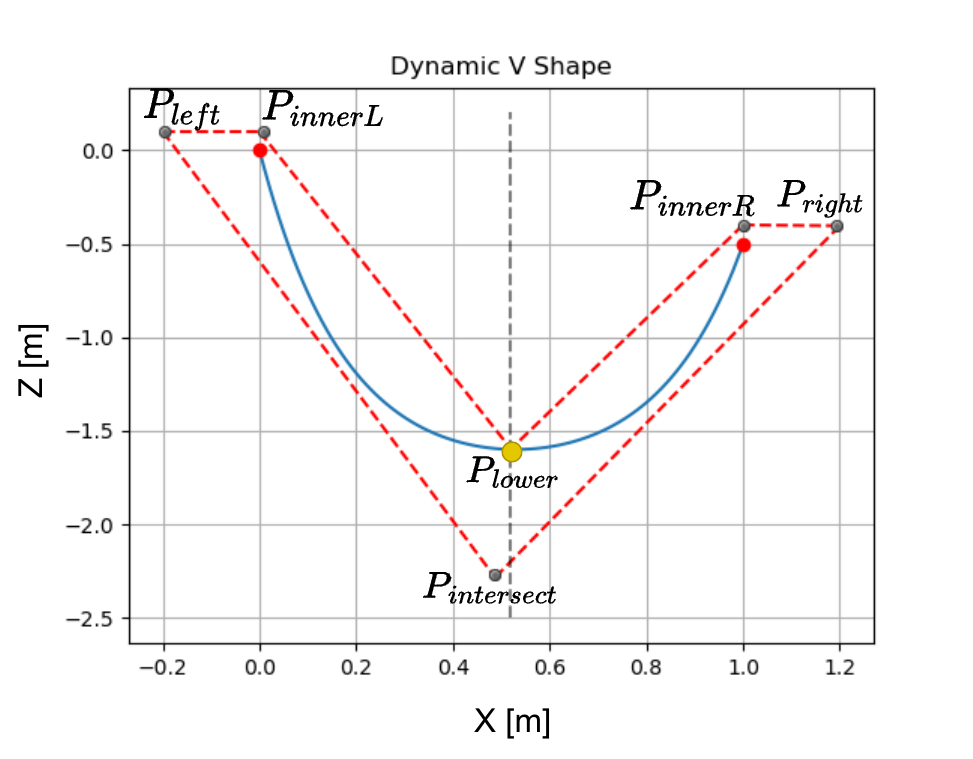} }} 
    \vspace{0.5cm}
    \qquad
    \subfloat[$3D$ Visualization]{{\includegraphics[width=6.7cm]{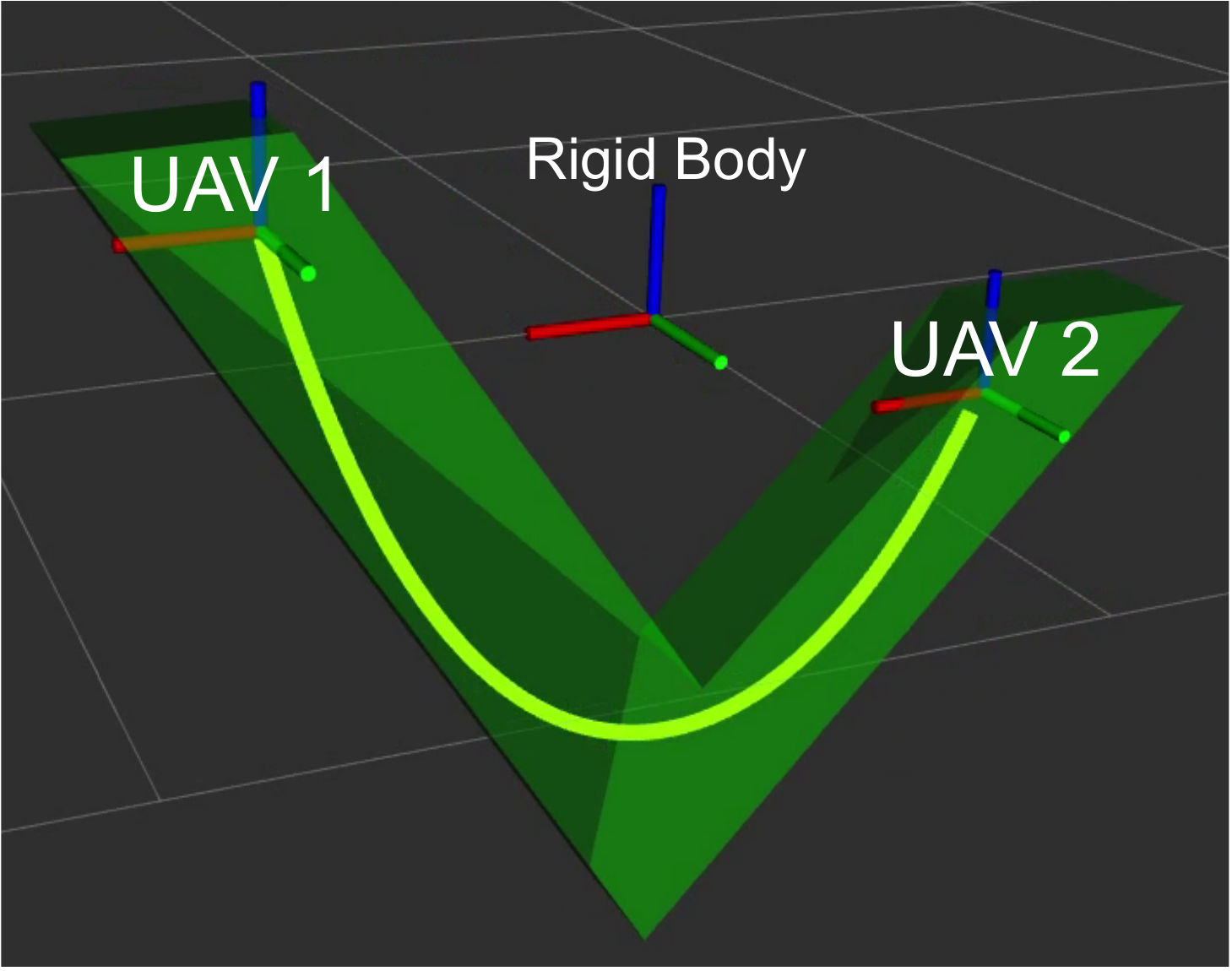} }}
    
    \caption{Dynamic V rigid body. The top figure (a) shows the initial calculation in the 2D plane and the bottom one (b) the final result in the 3D space enclosing the formation}
    \label{fig:dynamicV}
\end{center}
\end{figure}

The process to calculate it is depicted in Fig. \ref{fig:dynamicV}.

\begin{enumerate}
    \item $P_{lower} = [ argminf(x), minf(x) ]  $
    \item The left and right mounting points of the rope are considered to be  $\vec{P}_{innerR}$ and, $\vec{P}_{innerR}$ respectively.
    \item A safety horizontal offset is added to the right mounting point, $\vec{P_{right}} = \vec{P}_{innerR} + \vec{\Delta x}$, where $\vec{\Delta x} = [dx, 0]$
    \item Similarly for the left one, $\vec{P_{left}} = \vec{P}_{innerL} - \vec{\Delta x}$
    \item  $\mathbf{C}= \{ P_{i} = f(x_i), i=o,\dots,n \}$ is the catenary curve points, where $x_i = P^x_{innerL} + i*dx , i=0,1,\dots,n $ with $n = ( {P^x_{innerR} - P^x_{innerL} }) / {dx} $
    \item In order to find the tangent lines to the catenary curve passing through $P_{left}$ and $P_{right}$, a point $\mathbf{T_{right}} \in \mathbb{R}^2 = [P^x_{lower}, y] : det(M_i)>0 $ is calculated, where $i= 0,1,\dots,n$ and $M_i = \begin{bmatrix}
        \vec{P_{right}} - \vec{T_{right}} &
        \vec{C_i} - \vec{T_{right}}
        \end{bmatrix}, C_i \in \mathbf{C}$.
    The line ${l_{right}}$ connecting the points $\vec{P_{right}}$ and $\vec{T_{right}}$ splits the space in two halves, with the upper one containing all the curve points $C_i \in \mathbf{C}$.
    \item A similar procedure is followed to get  point $\vec{T_{left}}$ by using $\vec{P_{left}}$ instead of $\vec{P_{right}}$ and then calculate line ${l_{left}}$.
    \item The point $P_{intersect}$ of intersection of lines ${l_{left}}$ and ${l_{right}}$ is calculated.
    \item The transformation of the shape in the $3D$ space is executed by using the same points padded with an offset in the $Z$-axis perpendicular to the $2D$ plane in Fig. \ref{fig:dynamicV}.b, this offset represents the thickness of the shape.
\end{enumerate}

A 3D visualization of the dynamic rigid body is available at \href{https://youtu.be/IXeX2oimCeI}{https://youtu.be/IXeX2oimCeI}.

\subsection{State Validity} \label{subsec:state_validity}
Each state $X\in \mathbb{R}^6$ of the formation is expressed in Eq.~\ref{eq:planning_state}.

\begin{equation}\label{eq:planning_state}
    X=[x,y,z,\phi_{yaw},d,\theta_{form}]
\end{equation}

\begin{figure}[ht]
\begin{center}
\includegraphics[width=\columnwidth]{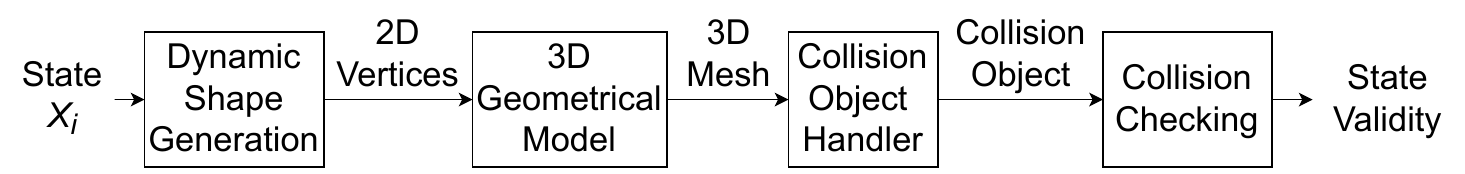}    
\caption{State Validity Check} 
\label{fig:state_validity_check}
\end{center}
\end{figure}

The planning state space is a 6-dimensional space that is bounded by the physical limitations of the formation, like the length of the rope and the dimensions of the space, where the planning takes place. The validation of each state generated while planning must be tested. The process for doing that is shown in Fig. \ref{fig:state_validity_check}, and is presenting in the following steps:

\begin{enumerate}
    \item The state $X_i$ is generated to be tested for validity.
    \item The Dynamic shape generation mechanism calculates the vertices of the dynamic $V$ shape.
    \item The new vertices are fed to the geometrical model that represents the formation. Its collision object is updated based on them.
    \item Collision checking is executed between the formation and the pre-loaded environment collision objects. The outcome of the collision checking determines the validity of the state.
\end{enumerate}

Both environment obstacles (which are given as Standard Triangle Language files, .stl) and dynamic rigid body are represented by Bounding Volume Hierarchy (BVH) models and Flexible Collision Library \cite{fcl} was chosen as the engine responsible for collision checking.

Due to the high dimensionality of the search space, the path planning algorithm was based on Rapidly Exploring Random Tree (RRT)~\cite{rrt_paper} and as it will be presented, it has the tendency to expand towards large unsearched areas, and it has been shown to be more efficient than the other traditional algorithms (Dijkstra, A*, etc.) in higher dimensions \cite{rrt-a*-comparison}. The path planning was implemented through the Open Motion Planning Library \cite{sucan2012the-open-motion-planning-library} framework, which provided an interface for setting up the problem configuration and handling the planning process. As soon as the path is extracted, it is simplified by applying short-cutting, vertices reduction and close vertices collapsing. After having the final formation path, decomposition is carried out by implementing the inverse transform stated in Section \ref{sec:composition_decomposition} which leads to the two lists of waypoints each UAV has to go simultaneously.

\begin{figure}[htbp]
\begin{center}
\includegraphics[width=\columnwidth]{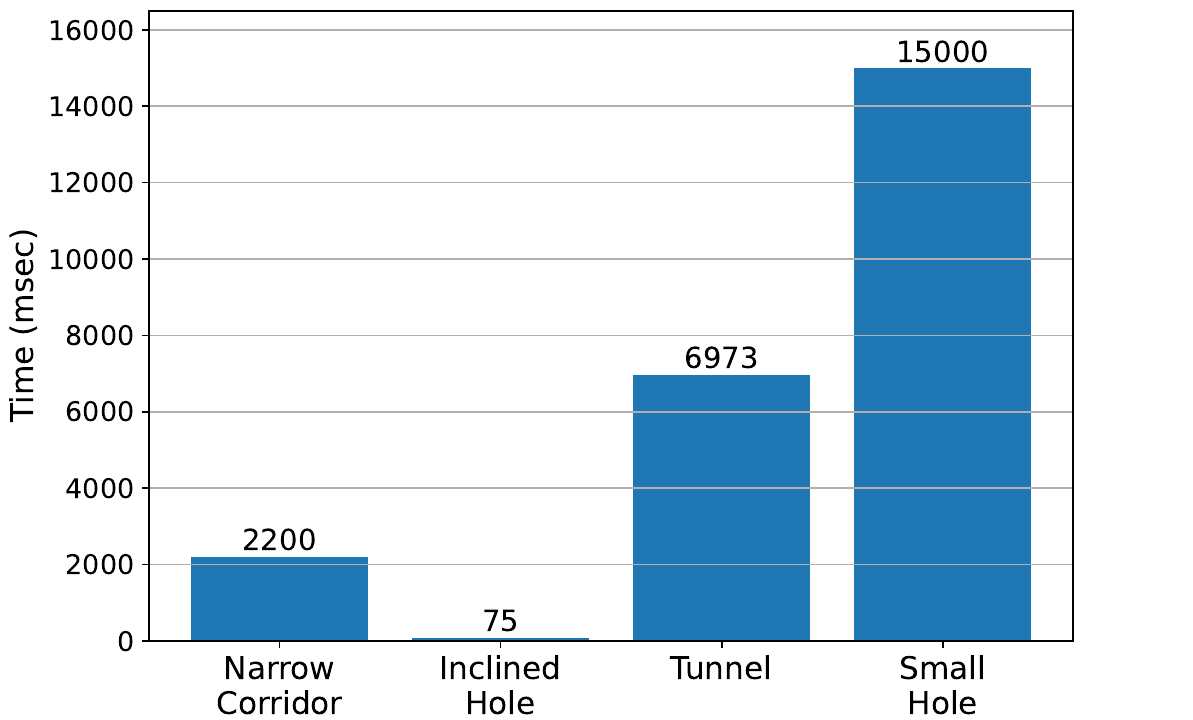}
\caption{Path planning execution times for different obstacles} 
\label{fig:execution_times}
\end{center}
\end{figure}

The path planning execution times for different obstacles can be shown in Fig. \ref{fig:execution_times}. Planning was carried out on a laptop with $Intel\textsuperscript{\textcopyright} \ Core\textsuperscript{\texttrademark} \ i5-7200U \ CPU \ @2.50GHz \ \times \ 4$ Processor and $8GB \ RAM$, while the overall time depends on the complexity of each object along with the number of degrees of freedom of the formation that must be utilized in order to avoid or pass through it.

\section{UAVs control Framework}
\subsection{Position Control}

A Model Predictive Control (MPC) scheme, based on \cite{bjorn_mpc}, was used for enabling collision avoidance and optimal behavior, since the system has been tested in constrained environments. MPC is a popular controller among researchers working with UAVs, due to its performance and predictive behavior. It optimizes a predefined cost function for each quadrotor, aiming at giving a position waypoint in space as a reference. A low-level attitude controller is being executed in the quadrotor and can track the desired roll and pitch $\phi_d,\theta_d$ angles and thrust $T$, that are generated by the MPC, with a first-order behavior with gains of $K_\phi, K_\theta \in \mathbb{R}^2$ and time constants of $\tau_\phi, \tau_\theta \in\mathbb{R}^2$. The magnitude and the angle of the thrust vector, produced by the motors, along with linear damping terms $A_x, A_y, A_z\in \mathbb{R} $ and the earth gravity $g\in \mathbb{R}$ are assumed as the only factors that affect acceleration in this particular model. The dynamic model of the system is described by Eq.~\ref{eq:quadcopter_dynamics}.


\begin{align}
&\dot{p}(t) = v_{z}(t) \nonumber\\
&\dot{v}(t) = R_{x,y}(\theta,\phi) \begin{bmatrix} 0\\ 0\\ T\end{bmatrix} + \begin{bmatrix} 0\\ 0\\ -g\end{bmatrix} - \begin{bmatrix} A_{x} & 0 & 0\\ 0 & A_{y} & 0\\ 0 & 0 & A_{z}\end{bmatrix}u(t) \label{eq:quadcopter_dynamics}\\
&\dot{\phi}(t) = \frac{1}{\tau_{\phi}} (K_{\phi} \phi_{ref}(t) - \phi(t)) \nonumber\\
&\dot{\theta}(t) = \frac{1}{\tau_{\theta}} (K_{\theta} \theta_{ref}(t) - \theta(t)) \nonumber
\end{align}

The vector $X=[p,u,\phi,\theta]$ is the state for each UAV and the vector $u=[T,\phi_{ref},\theta_{ref}]$ is the control input. Using the Forward Euler method, the quadrotor model is discretized with a sampling time $\delta_t \in \mathbb{Z}^+$ for each time instant $(k+j|k)$, which denotes the prediction of the time step $k+j$ produced at the time step $k$. The final goal of the controller is to navigate smoothly to the reference position, so the cost function is consisted of three terms.

\begin{equation}\label{eq:MPC_cost_func}
\begin{gathered}
J=\sum_{j=1}^{N}\underbrace{(x_{ref}-x_{k+j|k})^T Q_x (x_{ref}-x_{k+j|k})}_\text{state\space cost} \\
+\underbrace{(u_{ref}-u_{k+j|k})^T Q_u (u_{ref}-u_{k+j|k})}_\text{input\space cost}\\
+\underbrace{(u_{k+j|k}-u_{k+j-1|k})^T Q_{\Delta u} (u_{k+j|k}-u_{k+j-1|k})}_\text{input\space smoothness \space cost}
\end{gathered}
\end{equation}

$Q_x \in \mathbb{R}^{8\times8},Q_u,Q_{\Delta_u} \in \mathbb{R}^{3 \times 3} $ are weight matrices for the state weights, input weights and input rate weights respectively. The first term of Eq.~\ref{eq:MPC_cost_func} describes the state cost, which is the cost associated with deviating from a certain state reference $x_{ref}$. The second term describes the input cost that penalizes a deviation from the steady-state input, $u_{ref} = [g, 0, 0]$ i.e., the inputs that describe hovering. To guarantee the minimum control effort and smooth control actions, the third term is included, which compares the input at $k+j-1|k$ with the input at $k+j|k$ and penalizes the change of the input from one time step to the next one. It is finally minimized by using Optimization Engine (OpEn) Solver \cite{open2020}.

\subsection{Trajectory execution}
\begin{figure}[ht]
\begin{center}
\includegraphics[width=8.4cm]{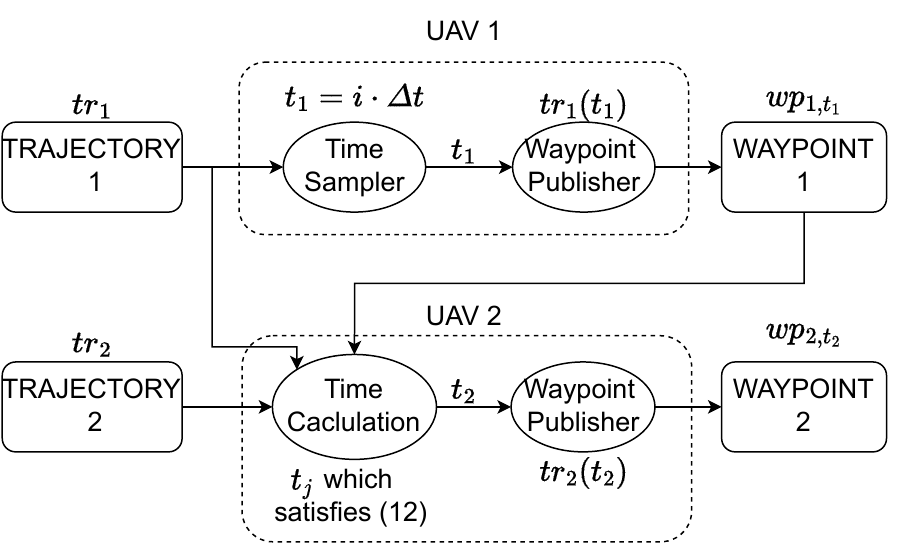}    
\caption{Synchronized Execution Mechanism} 
\label{fig:sync_exec_mechanism}
\end{center}
\end{figure}

A piecewise-polynomial trajectory $tr_i \in \mathbb{R}^3$, $i=1,2$ for each quadrotor is generated that passes through all the waypoints assigned to it. In order to guarantee the simultaneous execution of these two, the mechanism shown in Fig. \ref{fig:sync_exec_mechanism} is introduced.

The 2 UAVs are named UAV1 and UAV2 (order does not matter). UAV1 is only aware of its own piece-wise polynomial trajectory, $tr_{1}$ while UAV2 receives both trajectories $tr_{1}$ and $tr_{2}$. UAV2 also has another extra input which is the waypoint $wp_{1,t}=tr_{1}(t)$ published by UAV1 waypoint publisher at time $t$. When they both receive the signal to start executing the trajectories, UAV1 publishes the first waypoint and the following, while using the discretized time $t_{1} = i*t_{step} ,i \in[0,\frac{T_{traj}}{t_{step}}]$. UAV2 receives this waypoint and since the polynomial for $tr_{1}$ is saved, the time $t_{1}$ can be easily calculated by solving the polynomial for the given $wp_{1}(t)$. Since each segment of the polynomial is consisted of $7th$ degree polynomials, it leads to 3 sets (1 for each coordinate) of 7 possible roots $tx^{j}_{possible}, ty^{j}_{possible}, tz^{j}_{possible},\ j\in [0,7]$. Starting from the first segment and moving towards the next one, a time $t_{j,possible}$ is going to be calculated, which is the same for all the coordinates and follows the criteria of Eq.~\ref{eq:criteria}.

\begin{equation}
\label{eq:criteria}
\begin{cases}
               tx^j_{possible},ty^j_{possible},tz^j_{possible}>0\\
               tx^j_{possible}=ty^j_{possible}=tz^j_{possible}\\
               tx^j_{possible},ty^j_{possible},tz^j_{possible} \in [t_k,t_k+d_k)
            \end{cases}
\end{equation}

$t_k,d_k$ are the start and the duration of the $k$-th segment of the trajectory. As soon as the time root that satisfies the above is found, it is considered as $t_2$, the $wp_2(t_2)$ is calculated and fed to the MPC. In each iteration, the calculation does not start from the first segment but continues from the one that the previous $t_2$ was calculated from. In case the desired time is not found, it then starts from scratch. The procedure is continued as mentioned until the end of the trajectory execution.

\section{Experimental Results}
\begin{figure}[ht]
\begin{center}
\includegraphics[width=7.2cm]{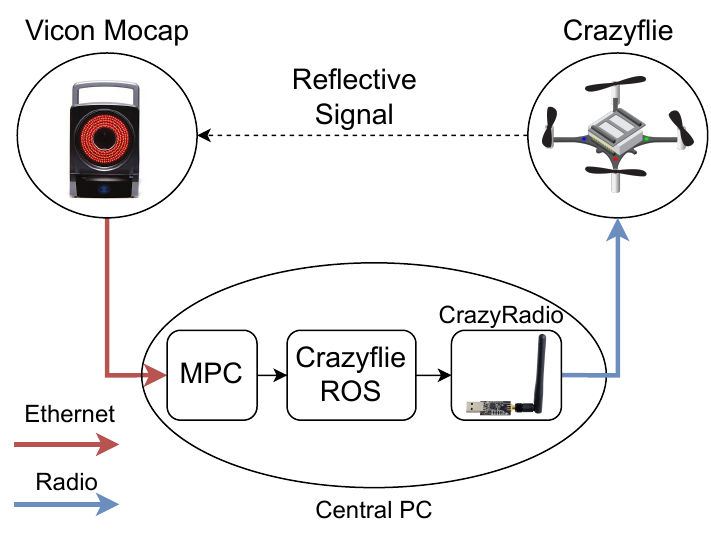}    
\caption{UAV Control System Architecture} 
\label{fig:UAV_control_diagram}
\end{center}
\end{figure}

For evaluating the efficiency of the proposed scheme, experiments were conducted in the flying arena of the Robotics and AI lab at Lule\aa\,\, University of Technology. The quadrotor used for conducting the experiments has been the Crazyflie 2.0.
A low-level controller implemented off-the-shelf is executed onboard, accepting desired attitude and thrust as input. All the path planning algorithms along with the MPC controller are executed in a central PC, which also receives the absolute position of each UAV through the Vicon motion capture system. The interface between all the sub-modules is achieved through ROS (Robot Operating System). The CrazyflieROS \cite{crazyflieROS} framework is utilized to communicate via radio with the Crazyflies. The whole architecture can be seen in Fig. \ref{fig:UAV_control_diagram}.

\begin{figure}[ht]
    \begin{center}
        \includegraphics[width=\columnwidth]{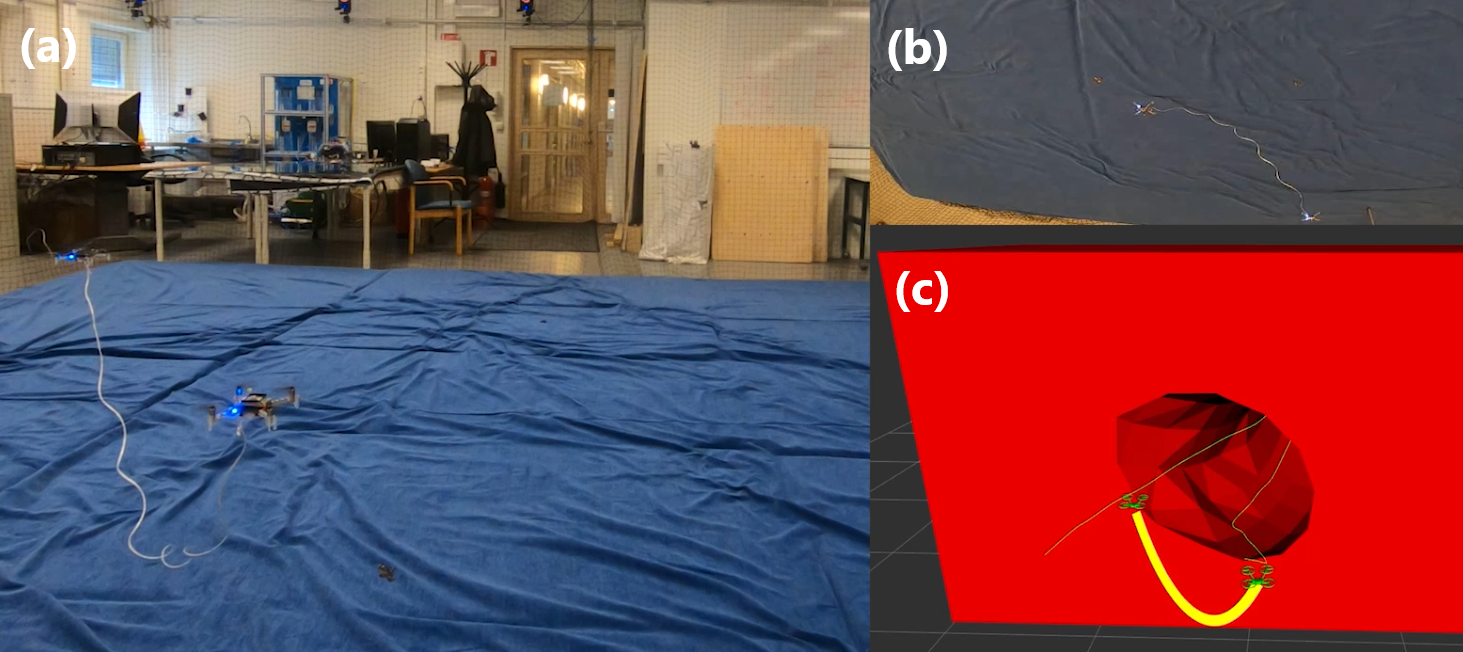}    
        \caption{Formation Path Planning through a tunnel (red).(a) Side View (b) Top
View (c) Visualization. Video available at \href{https://youtu.be/ApeidRGT8Kg}{https://youtu.be/ApeidRGT8Kg} }
        \label{fig:tunnel_experiment}
    \end{center}
\end{figure}

A snapshot of an experiment containing a tunnel-like obstacle can be seen in Fig. \ref{fig:tunnel_experiment}.  In this and the following experiment, there is no physical obstacle, but the real-time position of the UAVs provided by the motion capture system is visualized along with the red virtual obstacle, the rope shape and the paths generated for each UAV.

\begin{figure}[ht]
\begin{center}
\includegraphics[width=\columnwidth]{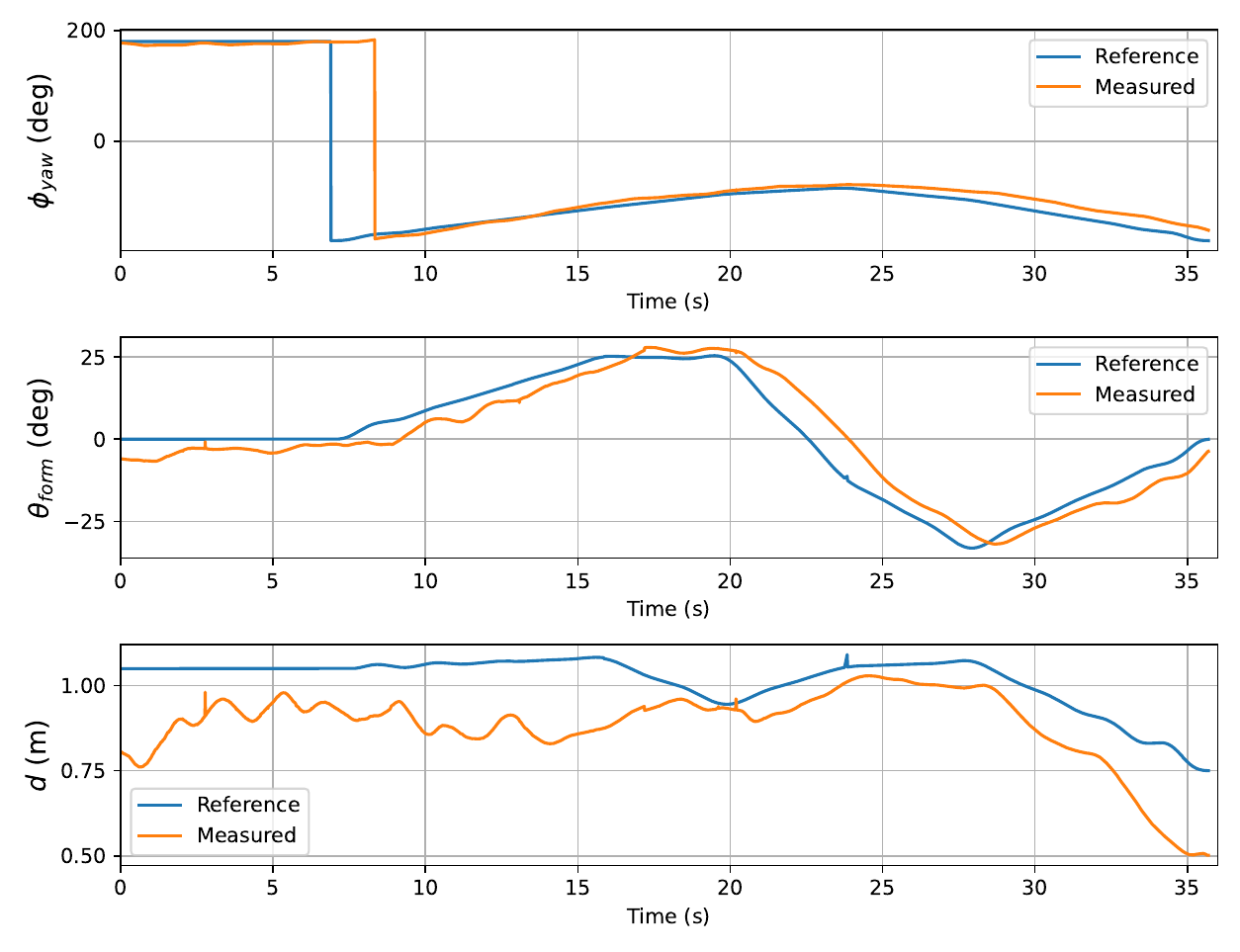}    
\caption{Formation's yaw, UAVs' angle $\theta$ and distance $d$ while passing through the tunnel} 
\label{fig:tunnel_form_chars}
\end{center}
\end{figure}

In Fig. \ref{fig:tunnel_form_chars} the desired (shown in blue) angles $\phi_{yaw}$, $\theta_{form}$ and distance $d$ along with their measured (shown in orange) values are shown. An error in the UAVs’ distance is noticed, but it is small enough not to cause any trouble in the path.

\begin{figure}[ht]
    \begin{center}
        \includegraphics[width=8.6cm]{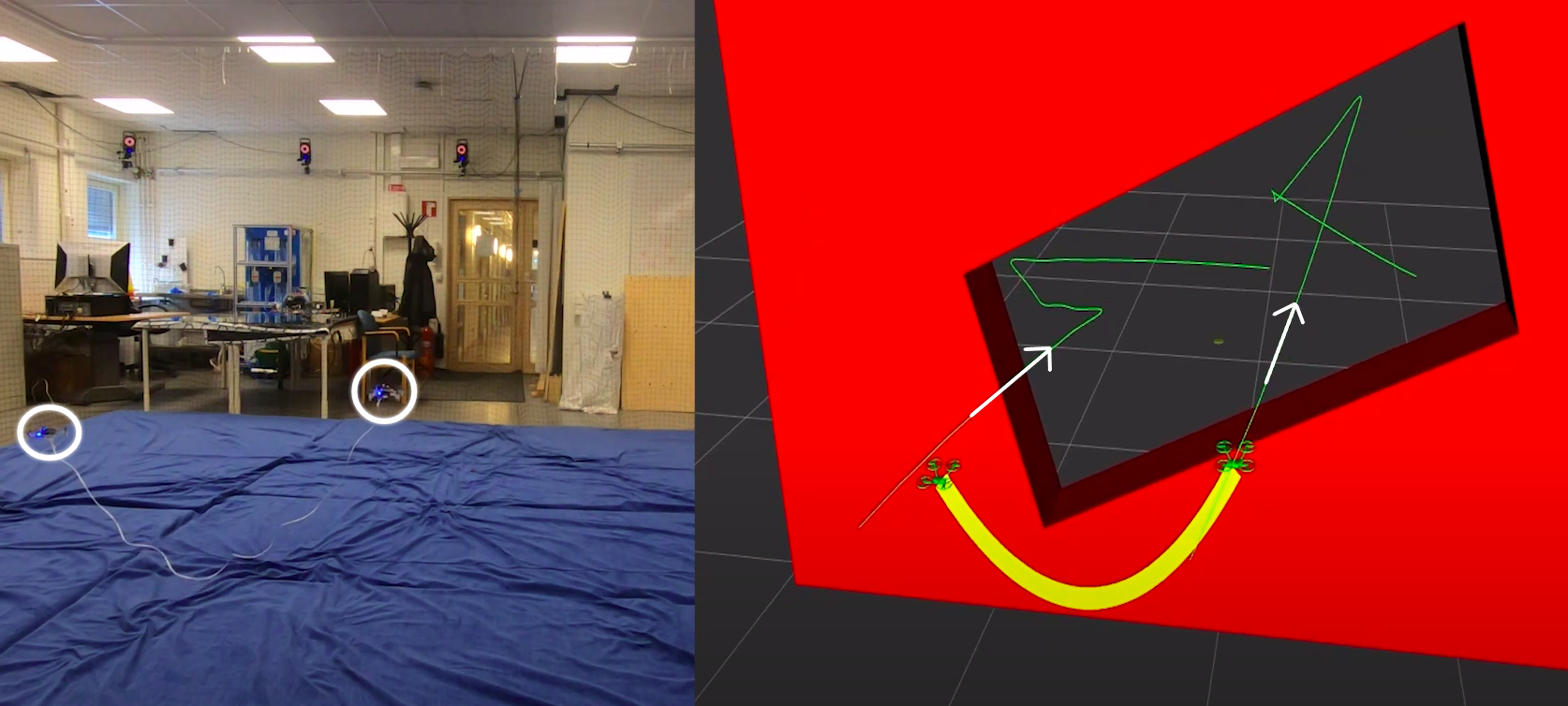}    
        \caption{Formation Path Planning through on inclined hall. (Left) Side View (Right) Visualization. Video available at \href{https://youtu.be/RSBxF--uA6g}{https://youtu.be/RSBxF--uA6g} }
        \label{fig:inclined_hole_experiment}
    \end{center}
\end{figure}

Another experiment is shown in Fig. \ref{fig:inclined_hole_experiment}. In similar fashion, the  obstacle demanded to pass through is an inclined parallelogram hole.  The dynamic formation needs to change its shape and adjust it to the shape of the whole, while increasing the distance between the UAVs to reduce the horizontal distance from the lowest point of the rope that would lead to possible collision otherwise.

\begin{figure}[ht]
\begin{center}
\includegraphics[width=\columnwidth]{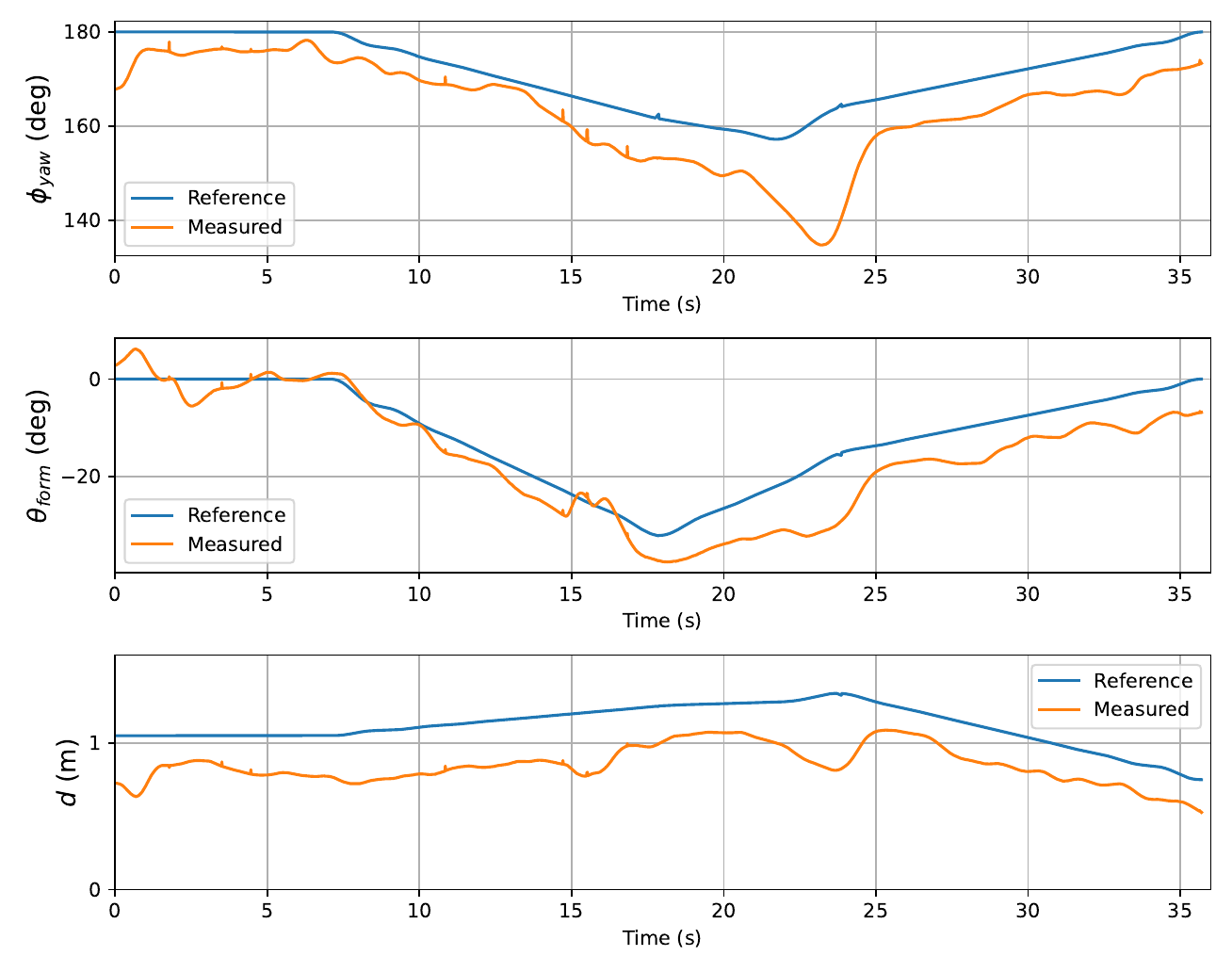}    
\caption{Formation's yaw, UAVs' angle $\theta$ and distance $d$ while passing through the inclined hole} 
\label{fig:inclined_hole_chars}
\end{center}
\end{figure}

The desired and measured formation features for the inclined hole experiment can be seen in Fig. \ref{fig:inclined_hole_chars}. A relatively big error on the yaw angle can be noticed at the 22nd second but is later compensated. An error on the UAVs distance $d$ is also noticed but due to the safety offsets introduced in \ref{subsec:state_validity} does not cause any collision with the obstacle.

\section{Conclusion}


In this work, a compact and simplified transformation was introduced to express the states of the formation of the two UAVs and the rope connecting them. In combination with the dynamic $V$ rigid body generation, a fast collision checking was implemented and led to the first path planning scheme of such kind of system in the literature. This could be applied to firefighting scenarios by transferring the fire hose into buildings and in search and rescue missions by delivering supplies to individuals trapped in confined environments like tunnels or mines. By utilizing the above, it is able to avoid obstacles and pass through them while keeping a successful lift. 

Integrating the current approach into dynamically changing environments along with a more precise physical model of the rope could be introduced in future work. Thus, a closer real use-case scenario execution would be made and the assumption of the rope being static and not swinging because of inertia would be no longer present.


\bibliographystyle{./IEEEtranBST/IEEEtran}
\bibliography{./IEEEtranBST/IEEEabrv,references}

\end{document}